\documentclass{article}
\usepackage{ijcai19}

\usepackage{times}
\usepackage[colorlinks]{hyperref}
\usepackage{soul}
\usepackage{url}
\usepackage[utf8]{inputenc}
\usepackage[small]{caption}
\usepackage{graphicx}
\usepackage{amsmath,amssymb}
\usepackage{mathtools}
\usepackage{booktabs}
\usepackage{algorithm}
\usepackage{algorithmic}
\newcommand{\methodName}{{CERTIFAI}}
\usepackage{tikz}

\newcount\Comments   
\usepackage{color}
\definecolor{purple}{rgb}{1,0,1}
\definecolor{orange}{rgb}{.85,.25,0}
\newcommand{\kibitz}[2]{\ifnum\Comments=1\textcolor{#1}{#2}\fi}

\title{\methodName: Counterfactual Explanations for Robustness, Transparency, Interpretability, and Fairness of Artificial Intelligence models}

\author{
Shubham Sharma$^1$
\and
Jette Henderson$^2$\and
Joydeep Ghosh$^2$
\affiliations
$^1$Department of Electrical and Computer Engineering, University of Texas at Austin\\
$^2$CognitiveScale\\
\emails
shubham\_sharma@utexas.edu,
jhenderson@cognitivescale.com,
jghosh@utexas.edu
}

\begin{document}

\maketitle

\begin{abstract}
As artificial intelligence plays an increasingly important role in our society, there are ethical and moral obligations for both businesses and researchers to ensure that their machine learning models are designed, deployed, and maintained responsibly.  
These models need to be rigorously audited for fairness, robustness, transparency, and interpretability.
A variety of methods have been developed that focus on these issues in isolation, however, managing these methods in conjunction with model development can be cumbersome and time-consuming. 
In this paper, we introduce a unified and model-agnostic approach to address these issues: Counterfactual Explanations for Robustness, Transparency, Interpretability, and Fairness of Artificial Intelligence models (CERTIFAI). Unlike previous methods in this domain, \methodName~is a general tool that can be applied to any black-box model and any type of input data. 
Given a model and an input instance, \methodName~uses a custom genetic algorithm to generate counterfactuals\footnote{ Counterfactuals have a well-established meaning in the causality literature. However, we are using  “counterfactual” in the counterfactual explanation sense, one that has been recently  initiated in  the explainability literature \cite{wachter2017counterfactual} where the model implies a machine learning model and not a causal model and hence no causal assumptions are made here}: instances close to the input that change the prediction of the model. 
We demonstrate how these counterfactuals can be used to examine issues of robustness, interpretability, transparency, and fairness. Additionally, we introduce CERScore, the first black-box model robustness score that performs comparably to methods that have access to model internals.

\end{abstract}

\section{Introduction}

As the adoption of machine learning models to everyday tasks grows rapidly, so does the need to consider the ethical, moral, and social consequences of the decisions made by such models.
Several important questions arise in a variety of applications, such as the following: 1) how did the model predict what it predicted?, 2) if a person got an unfavorable outcome from the model, what can they do to change that?, 3) has the model been unfair to a particular group?, and 4) how easily can the model be fooled? 
Researchers are actively building separate approaches to answer each of these questions.  

\begin{figure}[tp]
\centering
\includegraphics[width = 8.5cm]{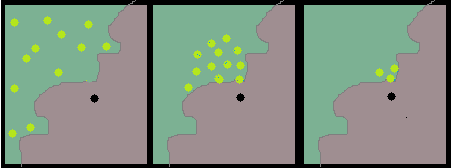}
\caption{The \methodName~counterfactual generation process. The decision boundary for a binary classifier is shown, with the input instance in black. We sample a set of points (left) in the feature space with a constraint that they must lie on the other side of the decision boundary (green points). The algorithm then evolves these samples (middle) to generate individuals that lie closer to the input point but on the other side of the decision boundary. Finally, a smaller set, the size of which is user-defined, of counterfactuals is generated (right).}
\label{fig:CE_example}
\end{figure}


\begin{table*}[ht]
\def\checkmark{\tikz\fill[scale=0.4](0,.35) -- (.25,0) -- (1,.7) -- (.25,.15) -- cycle;} 
\centering
\begin{tabular}{lrrrrrrrr}  
\toprule
Method & Black-box & Model-Agnostic & Mixed-data  &  Explainability & Fairness & Robustness\\
\midrule
CERTIFAI & \checkmark & \checkmark & \checkmark & \checkmark & \checkmark & \checkmark     \\
\cite{ustun2019actionable} & \checkmark &  & \checkmark & \checkmark & \checkmark &       \\
\cite{wachter2017counterfactual} & \checkmark & \checkmark &  & \checkmark & \checkmark &       \\
\cite{russell2019efficient} & \checkmark &  & \checkmark & \checkmark &\checkmark &       \\
\cite{ribeiro2016should} & \checkmark & \checkmark & \checkmark  & \checkmark &  &       \\
\cite{guidotti2018local} & \checkmark & \checkmark & \checkmark & \checkmark &  &       \\
\cite{carlini2017towards} &  & \checkmark &  &  &  & \checkmark       \\
\cite{weng2018evaluating} &  & \checkmark &  &  &  & \checkmark       \\
\bottomrule
\end{tabular}
\caption{Comparison of related work with our approach. We consider the approach most similar to ours. Mixed-data means the method can work with both discrete and continuous data, without any discretization or assumptions.}
\label{tab:relwork}
\end{table*}

One promising vein of research in explainability, first introduced by \cite{wachter2017counterfactual}, is generating counterfactuals. 
Given an input data point and a black-box machine learning model (i.e. we only have access to the model's prediction for any input), a \emph{counterfactual} is defined as a generated data point that is as close to the input data point as possible but for which the model gives a different outcome.
For example, if a user was denied a loan by a machine learning model, an example counterfactual explanation could be: ``Had your income been \$5000 greater per year and your credit score been 30 points higher, your loan would be approved."
\cite{wachter2017counterfactual} argue that counterfactuals are a way of explaining model results to users such that they can identify actionable ways of changing their behaviors to obtain favorable predictions.
In addition to providing counterfactuals for explainability, we show how counterfactual explanations can be used to audit fairness and robustness of a model.


As promising as the original method~\cite{wachter2017counterfactual} and subsequent methods of generating counterfactuals~\cite{ustun2019actionable,russell2019efficient} are, they are also limited in that some only work for linear models, while others cannot deal with different data types.
To resolve these limitations, we introduce \methodName, a novel, flexible, model-agnostic technique for generating counterfactuals via a custom genetic algorithm.
The meta-heuristic evolutionary algorithm starts by generating a random set of points such that they do not have the same prediction as the input point.
A subsequent evolutionary process results in a set of points close to the input that maintain the prediction constraint. 
Figure~\ref{fig:CE_example} shows an example of three counterfactuals (green points) generated for a given input (black point). 


A major advantage of using the genetic algorithm to generate counterfactuals is that it can generate counterfactuals for linear and non-linear models (e.g. deep networks) and for any input form (from mixed tabular data to image data) without any approximations to or assumptions for the model.
Moreover, end-users can 1) define a range for any feature, and 2) restrict the features that can change. 
\methodName~simply constrains the values of the sampled points based on those choices, allowing the generated counterfactuals to reflect a user's understanding of how much it is \emph{possible} for them to change their features.


\methodName~can be used to audit any black-box classifier, providing three tools that are based on a single underlying algorithm. The major contributions of this paper are summarized as follows:
\begin{itemize}
\item Counterfactuals are generated using a custom genetic algorithm, which is model-agnostic, flexible, and can be used to provide explanations.
\item Counterfactuals are shown to be effective adversarial examples. They are also used to generate the Counterfactual Explanation-based robustness scores (CERScore), which to the best of our knowledge is the first ever black-box model robustness score.
\item Counterfactuals can be used to evaluate fairness with respect to a user's input as well as the fairness of the model towards groups of individuals.
\end{itemize}

\section{Related Work}

Table~\ref{tab:relwork} summarizes the key features of \methodName~and the work most related to~\methodName. Other general methods on explainability, fairness and robustness have been described by \cite{guidotti2018survey},\cite{binns2017fairness}, and \cite{akhtar2018threat} respectively. As we can see, there is no prior art that handles a diverse set of desirable properties needed to develop a responsible AI system. \cite{wachter2017counterfactual} introduced counterfactual explanations, however, their optimization formulation cannot handle categorical data (i.e. the optimization does not solve for such data; the values are found in a brute-force fashion). 
Methods in \cite{ustun2019actionable} and \cite{russell2019efficient} only work for linear models. \cite{guidotti2018local} uses a genetic algorithm to generate neighbors around the input and then use decision trees to locally approximate the model.  
However, local approximations might be at the cost of model accuracy, and they define counterfactuals based on the minimum number of splits in the trained decision tree, which might not always be the smallest change to the input. 
The closest work to generating adversarial examples is by \cite{carlini2017towards}. 
They work on a white-box and simpler convolution models, and the distance metrics might not be apt to measure image similarity. \cite{wang2002image}. 
\cite{weng2018evaluating} define a robustness score CLEVER. However, they have access to the model gradients. 

\section{The CERTIFAI framework}

In this section, we formulate a custom genetic algorithm to find counterfactual(s). 
Consider a black-box classifier \textit{f} and an input instance \textbf{x}. Let the counterfactual be a feasible generated point \textbf{c}.
Then the problem can be formulated as:

\begin{equation}
 \begin{aligned}
\min_{c} d(\textbf x,\textbf c)\\
\textrm{s.t.} f(\textbf c)\neq f(\textbf x)    \\
\end{aligned}
\label{eq:opt}
\end{equation}

\noindent where $d(\textbf x,\textbf c)$ is the distance between $\boldsymbol{x}$ and $\boldsymbol{c}$. To avoid using any approximations to or assumptions for the model, we use a genetic algorithm to solve Equation~\ref{eq:opt}. 
The custom genetic algorithm works for any black-box model and input data type, and it is model-agnostic.
Additionally, it provides a great deal of flexibility in counterfactual generation.

\methodName's genetic algorithm solves the optimization problem in Equation~\ref{eq:opt} through a process of natural selection. 
The only mandatory inputs for the genetic algorithm are the black-box classifier $f$ and an input instance \textbf{x}. 
Generally, for an $n$-dimensional input vector \textbf{x}, let $W\in \mathbb{R}^{n}$ represent the space from which individuals can be generated and $P$ be the set of points with the same prediction as \textbf{x}:
\begin{equation}
P = \{\textbf{p} |f(\textbf{p}) = f(\textbf{x}), \textbf{p} \in W\}.
\end{equation}
The possible set of individuals $\textbf{c}\in I$ are defined such that 
\begin{equation}
    I =  W\setminus P.\
\label{eq:I}
\end{equation}
Each individual $\textbf{c}\in I$ is a candidate counterfactual.
The goal is to find the fittest possible $\boldsymbol{c^*}$ to \textbf{x} constrained on  $\boldsymbol{c^*} \in I$.
The fitness for an individual \textbf{c} is defined as:
\begin{equation}
 \begin{aligned}
 fitness = \frac{1}{d(\textbf x, \textbf c)} .\\
\end{aligned}
\label{eq:fit}
\end{equation}
Here $\boldsymbol{c^*}$  will then be the point closest to \textbf{x} such that $\boldsymbol{c^*} \in I$.
For a multi-class case, if a user wants the counterfactual \textbf{c} to be belong to a particular class $j$, we define $Q$ as:

\begin{equation}
Q = \{\textbf{q} |f(\textbf{q}) = j, \textbf{q} \in W\}  .  
\end{equation}
Then Equation \ref{eq:I} becomes:
\begin{equation}
    I =  (W\setminus P) \cap Q.
\label{eq:Imod}
\end{equation}

The algorithm is carried out as follows:
first, a set $I_c$ is built by randomly generating points such that they belong to $I$. 
Individuals $\boldsymbol{c}\in I_c$ are then evolved through three processes: selection, mutation, and crossovers.
Selection chooses individuals that have the best fitness scores (Equation~\ref{eq:fit}).
A proportion of these individuals (dependent on $p_m$, the probability of mutation) are then subjected to mutation, which involves arbitrarily changing some feature values. 
A proportion of individuals (dependent on $p_c$, the probability of crossover) are then subjected to crossover, which involves randomly interchanging some feature values between individuals.
The population is then restricted to the individuals that meet the required constraint (Equation~\ref{eq:I} or Equation~\ref{eq:Imod}), and the fitness scores of the new individuals are calculated. 
This is repeated until the maximum number of generations is reached. 
Finally, the individual(s) $\boldsymbol{c^*}$ with the best fitness score(s) is/are chosen as the desired counterfactual(s)\footnote{ p$_m$=0.2 and p$_c$=0.5, which is standard in literature. The population size is the square of the input feature size with a maximum cap of 30,000. Grid-search is used to find the number of generations}.

\subsection{Choice of distance function}
The choice of distance function used in Equation~\ref{eq:opt} depends on the details provided by the model creator and the type of data being considered. 
If the data is tabular, \cite{wachter2017counterfactual} demonstrated how the $L_1$ norm normalized by the median absolute deviation (MAD) is better than using the $L_1$ or $L_2$ norm for counterfactual generation. 
For tabular data, the $L_1$ norm for continuous features (NormAbs) and a simple matching distance for categorical features (SimpMat) are chosen as default. In the absence of training data, normalization using MAD is not possible.
However in model development and our experimenets where there is access to training data, normalization is possible. The distance metric used is:
    
    \begin{equation}
 \begin{aligned}
  d(\textbf{x},\textbf{c}) = \frac{n_{con}}{n}\text{NormAbs}(\textbf{x},\textbf{c}) + \frac{n_{cat}}{n}\text{SimpMat}(\textbf{x},\textbf{c})\\
\end{aligned}
\end{equation}
\noindent where  $n_{con}$ and $n_{cat}$ are the number of continuous and categorical features, respectively, and $n$ is the total number of features ($n_{con} + n_{cat} = n$).  

For image data, the Euclidean distance and absolute distance between two images are not good measures of image similarity \cite{wang2002image}. 
Hence, we use SSIM (Structural Similarity Index Measure) \cite{wang2003multiscale}, which has been shown to be a better measure of what humans consider to be similar images \cite{wang2002image}.
SSIM values lie between 0 and 1, where a higher SSIM value means that two images look more similar to each other. For the input image \textbf{x} and counterfactual image \textbf{c}, the distance is: 
    
\begin{equation}
  d(\textbf{x},\textbf{c}) = \frac{1}{\text{SSIM}(\textbf{x},\textbf{c})}.
  \label{SSIMeq}
  \end{equation}
  
\subsection {Improving counterfactuals with constraints}

Apart from the input instance and black-box model, additional inputs help the algorithm produce better results. Auxiliary constraints are incorporated by restricting the space defined by the set $W$: the space from which individuals can be generated, to ensure feasible solutions. For an $n$-dimensional input, let $W$ be the Cartesian product of the sets $W_1$,$W_2$,...,$W_n$.
For continuous features, $W_i$ can be constrained as $W_i \in [W_{imin}, W_{imax}]$, and categorical features can be constrained as $W_i \in \{W_1, W_2,...,W_j\}$. However, certain variables might be immutable (e.g., race). In these cases, a feature $i$ for an input \textbf{x} can be muted by setting $W_{i}= x_{i}$.

An example of the benefits of such constraints would be when a user may not want an explanation of an income change from \$10,000 to \$900,000 if that is not possible, so $W_i \in [\$10000,\$15000]$ might be an appropriate constraint.  
The number of counterfactuals $k$ can also be set. 
\methodName~ chooses the top $k$ individuals ($k=1$ as default) where different features have changed, so the end-user can get multiple diverse explanations of different kinds.

\subsection {Robustness}
\label{sec:theory_rob}
Machine learning models are prone to attacks and threats. For example, deep learning models have performed exceedingly well for image recognition tasks, but it has been widely shown \cite{carlini2017towards}, \cite{nguyen2015deep} that these networks are prone to adversarial attacks. Two images may look the same to a human, but when presented to a model, they can produce different outcomes. A counterfactual is a generated point close to an input that changes the prediction and is therefore an adversarial example.

Given two black-box models, if the counterfactuals across classes are farther away from the input instances on average for one network as compared to the other network, that network would be harder to fool. Since \methodName~directly gives a measure of distance d(\textbf{x},\textbf{c}), this can be used to define the robustness score for a classifier.
Using this distance, we introduce Counterfactual Explanation-based Robustness Score (CERScore), the first ever black-box model robustness score. Given a model, the CERScore is defined as the expected distance between the input instances and their corresponding counterfactuals:

\begin{equation}
 \begin{aligned}
  CERScore(model)=  \mathop{\mathbb{E}}_{X}[d(\textbf{x},\textbf{c}^*)].
\end{aligned}
\label{eq:burd}
\end{equation}

To be able to better compare models trained on different data sets, the CERScore can be normalized by the expected value of the distance between data points in each class over all classes \textit{k}, and hence we get the normalized CERScore NCERScore (abbreviated as NC) as:
\begin{equation}
 \begin{aligned}
  NC=\frac{\mathop{\mathbb{E}}_{X}[d(\textbf{x},\textbf{c}^*)].}{\sum_{k=1}^{K} P(x\in \text{class}_k) \mathbb{E}[d(x_i,x_j);x_i,x_j \in \text{class}_k]}
\end{aligned}
\label{eq:cernorm}
\end{equation}
\noindent (i.e., we normalize by dividing by the expected distance between two datapoints drawn from the same class). A higher CERScore implies that the model is more robust. Note that the normalized CERScore can be greater than 1. Unlike \cite{weng2018evaluating}, \methodName~only needs model predictions and not the model internals.

\subsection{Fairness}
\label{sec:fair}
The fitness measure (Equation \ref{eq:fit}) and CERScore can also be used to investigate fairness from individual and group perspectives, respectively.
For a given individual instance, if the genetic algorithm can generate different counterfactuals with different values of a protected feature (e.g., race, age), and as a result the user can achieve the desired outcome more easily than when those features could not be changed, then the individual could claim the model is unfair to their case. 
Additionally, \methodName~can be used by model developers to audit the fairness for different groups of observations.
If the fitness measure is markedly different for counterfactuals generated for the different partitions of a feature's domain value, this could be an indication the model is biased towards one of the partitions.
For example, if the gender feature is partitioned into two values (male and female), and the average fitness values of generated counterfactuals are lower for females than for males, this could be used as evidence that the model is not treating females fairly.
Using counterfactuals and the distance function, we can calculate the overall burden for a group, measured as:
\begin{equation}
 \begin{aligned}
  Burden(g)=  \mathop{\mathbb{E}}_{g}[d(\textbf{x},\textbf{c}^*)]
\end{aligned}
\label{eq:burd}
\end{equation}
where \emph{g} is a partition defined by the distinct values for a specified feature set. 
Note, burden is related to CERScore as it is the expected value over a group. 
Most fairness auditing models focus on single features (e.g., \cite{hardt2016equality,donini2018empirical}).  Burden, however, does not have that limitation and can be applied to any combination of features.

\section{Experiments}

\begin{figure}[tp]
\centering
\includegraphics[width = 6cm, height = 4cm]{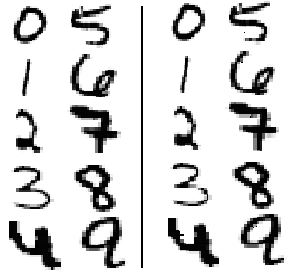}
\caption{Adversarial examples for the MNIST dataset. The images on the left are original inputs for which the model has a correct prediction and the images on the right are counterfactual images (adversarial examples) for which the prediction has changed.}
\label{fig:adv_examples}
\end{figure}
We demonstrate the applications and flexibility of \methodName~ to explainability, transparency, fairness, and robustness.

\subsection{Robustness}
In this section, we demonstrate how \methodName~ produces adversarial examples, and we use CERScore from Section~\ref{sec:theory_rob} to measure a network's resistance to adversarial attacks.

\subsubsection{Generating Adersarial Examples}
We consider the MNIST dataset \cite{lecun1998mnist} which contains 60000 (size 28x28) training images of digits. We use it to train a convolutional neural network, consisting of one convolution layer, two dense layers, and intermediate pooling and dropout layers. We achieve a 99.46\% accuracy on the test set using this architecture. Then we select random images from the MNIST dataset and the model above and find the counterfactual image for each image, using SSIM (equation (8)). Every pixel is considered to be a feature and hence, every individual in the population is a 784 dimension vector. We use an initial population size of 30,000 individuals and run the experiment for 1,000 generations. 

\begin{table}
\centering
\begin{tabular}{lrrrr}  
\toprule
Model & CERScore & CI &CLEVER\\
\midrule
Inception-v3   & 1.17 & 1.09-1.25 &0.229 \\
Resnet-50  &        1.06  & 1.05-1.08 &0.137     \\
MobileNet &     1.08    & 1.06-1.09 &0.151        \\
\bottomrule
\end{tabular}
\label{tab:CERScore}
\caption{Robustness score and 95 percent confidence intervals (CI) for those scores for 3 deep learning models and the corresponding CLEVER scores} 
\end{table} 

\begin{table*}[]
\centering
\begin{tabular}{lcccccccc}
\toprule
\textbf{Data set} & \textbf{Num.}           & \textbf{Num.}  & \multicolumn{2}{c}{\textbf{DT}}  & \multicolumn{2}{c}{\textbf{SVM}}   & \multicolumn{2}{c}{\textbf{MLP}}  \\ 
                 & \textbf{obs.} & \multicolumn{1}{l}{\textbf{features} } & \textbf{NCERS.} & \textbf{Acc.} & \textbf{NCERS.} & \textbf{Acc.} & \textbf{NCERS.} & \textbf{Acc.} \\
                 \midrule
Pima Diabetes         & 768                  & 8                    &       0.074            &      73.25         &     0.387              &        81.42       &  \textbf{0.486}                 & 98.61              \\
Breast Cancer   & 569                  & 32                   &           0.081        &      95.80         &         \textbf{0.121}          &    96.50           &  \textbf{0.124}                 &      96.50         \\
Iris   \small{ }           & 150                  & 4                    &    0.132               &       95.67        & 0.235                  &        95.67       &       \textbf{0.241}            &  95.67              \\ \bottomrule
\end{tabular}
\label{tab:data}
\caption{Descriptions of data sets, and NCERScore (NCERS.) and test set accuracy (Acc.) for three models: decision tree (DT), SVM with RBF kernel (SVM), and Multilayer Perceptron (MLP).}
\end{table*}
Figure~\ref{fig:adv_examples} shows an example of ten generated counterfactuals (right) and their original counterpart (left) images. 
The counterfactual images on the right look nearly identical to the input images on the left, however, the model predicts a different outcome for the images on the left and right. The imperceptibly different images give credence to the idea of using a genetic algorithm formulation to produce counterfactuals. Additionally, our approach towards generating these images is model-agnostic and does not require any approximations, unlike \cite{carlini2017towards}. The generated images show how a network can easily be fooled and demonstrate that there is a major problem in deploying such highly-accurate networks to image-based decision making applications (eg. face recognition). Moreover, different kinds of adversarial attacks can be generated by simply changing the distance function in Equation \ref{eq:fit}.
\subsubsection{Evaluating Deep Networks}

In this section, we evaluate how well CERScore, introduced in Section~\ref{sec:theory_rob}, can give an informative measure of robustness.
We consider the same networks as in \cite{weng2018evaluating}: Inception-v3 \cite{szegedy2016rethinking} , ResNet-50 \cite{he2016deep} and MobileNet \cite{howard2017mobilenets} pre-trained on ImageNet \cite{deng2009imagenet}, where they define the CLEVER score for robustness. Unlike CLEVER, we consider the model to be a black-box (only relying on its predictions). Ideally, to derive a measure of robustness for a model, all images from all classes should be considered, their counterfactuals should be generated, and the CERScore should then be calculated. However, since the number of training samples for a deep network is in the order of millions, it is not computationally feasible to calculate the score for each example. Hence, we consider a subset of classes and images to calculate the CERScore. We sampled \emph{n}=50 random images from every class across \emph{k}=100 random classes. We generate the counterfactuals for all 5,000 images such that the counterfactual gives a prediction of the second most likely class (by generating individuals constrained on belonging to that class as in Equation~\ref{eq:Imod}) and empirically estimate the CERScore as:

\begin{equation}
 \begin{aligned}
  CERScore= \frac{1}{nk}\sum_{i=1}^k \sum_{j=1}^n d(\textbf{x$_i$$_j$},\textbf{c$^*_{ij}$})
\end{aligned}
\end{equation}
\noindent where $\textbf{x}_{ij}$ is the $j^{th}$ input instance belonging to predicted class $i$, and $\textbf{c}^*_{ij}$ is the corresponding counterfactual. The CERScores are shown in Table~\ref{tab:CERScore}. One way to interpret the score is that on average, the SSIM score for Inception-v3 is 1/1.17 = 0.85, where an SSIM score of 1 means the images look exactly the same and an SSIM score of 0 means the images are highly different. Hence, adversarial attacks for Inception-v3 could be more easily identified than for the other models. 
We also show the 95\% confidence interval where we have assumed the distribution of distances between the images and their counterfactuals follows a normal distribution.
The confidence intervals are tight around the CERScores.

We also compare CERScore with CLEVER scores \cite{weng2018evaluating} for the same images, considering the top-2 class attack. The CLEVER scores are also reported in Table~\ref{tab:CERScore}. The CERScore implies that Inception-v3 is most robust and Resnet-50 is least robust, which is similar to what the CLEVER scores suggest. Hence, even though \methodName~ does not access any model weights, it is able to evaluate a model's robustness to adversarial attacks.
\subsubsection{Robustness of Classic Classifiers}

Next, we use NCERScore (Equation \ref{eq:cernorm}) to compare the robustness of different models trained on different data sets.
We train three models (decision trees (DT), Support Vector Machines with RBF kernel (SVM), and multilayer perceptrons (MLP)) on the three data sets listed in Table~\ref{tab:data}.
We report the NCERScore and the accuracy on the test set in Table~\ref{tab:data}.
Across all data sets, the neural network has the highest NCERScore and is therefore the most robust of the classifiers for these data sets. 
In the Pima diabetes data set, the accuracy of the decision tree is much lower than the other models, which suggests this simple model cannot adequately capture the class separation.
Hence, more points would be concentrated near the decision boundaries, resulting in a lower NCERScore. 

\begin{figure*}[tp]
\centering
\includegraphics[width = 14cm, height = 5cm]{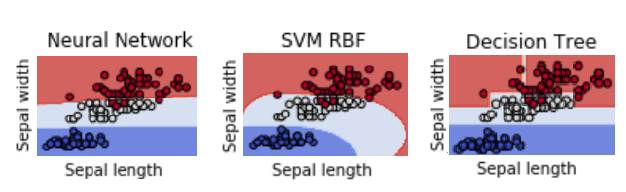}
\caption{Decision boundaries and input points for the Iris flowers data set (3 classes) for 3 models: Neural Network, SVM with RBF kernel and Decision tree (DT), visualized using two features from the data set. DT has closer points to the decision boundaries on average.}
\label{fig:DT}
\end{figure*}

For the Iris data set, while it is a relatively simple data set (even the decision tree performs well), the decision tree has the lowest NCERScore while the scores for SVM and MLP are similar. 
In Figure~\ref{fig:DT}, we plot input points for two features of the Iris data set and the decision boundary for each model. 
Looking closely, the points for the decision tree are closer on average to the decision boundaries as compared to the other two models (i.e. the densely clustered white and red points are closer to decision boundaries), which suggests the model is more prone to being fooled. 
The decision boundaries for both SVM and MLP are nearly identical around the input points, which results in similar robustness scores. 
Similar results can be seen for the cancer data set. Using these results, a model developer can choose the most robust model based on the NCERScore.
\begin{table}
\centering
\begin{tabular}{lrrrr}  
\toprule
Person & Feature(s) & Original & Counterfactual\\
\midrule
1 & Glucose (CWC)       & 115  & 71     \\
    &BMI   (CUC)         & 35.3  & 10.1       \\
\midrule
2 & Glucose (CWC)       & 168  & 89     \\
    &Age    (CUC)        & 34  & 44       \\
\bottomrule
\label{tab:PIMA}
\end{tabular}
\caption{Counterfactual explanations for the Pima Indian diabetes dataset. CWC: counterfactuals with constraints on feature values and CUC: unconstrained counterfactuals. Unconstrained features lead to infeasible solutions (BMI 10.1) or unchangeable features (age) being changed.}
\label{tab:Hi}
\end{table}

\subsection{Explainability}
\begin{table}
\centering
\begin{tabular}{lrrrr}  
\toprule
Person & Feature(s) & Original & Counterfactual\\
\midrule
1 & Education       & 12th  & Bachelors     \\
    & Occupation         & Tech-suppt  & Exec-managerial       \\
\midrule
1 & Hrs-per-week       & 50  & 70     \\
    &Workclass        & Local-gov  & Private       \\
\bottomrule
\label{tab:UCI}
\end{tabular}
\caption{Two explanations for the same person from the UCI adult dataset, with constraints on feature values.}
\label{tab:boo}
\end{table}
Counterfactuals are used to provide explanations and transparency to a user on how much change is needed for them to obtain a favorable prediction. We show the importance of using the constraints to improve explanations, the use of multiple counterfactual explanations for a single instance, and how these can also be used to estimate feature importance.
\subsubsection{Datasets and Models}
We use the Pima Indian dataset \cite{smith1988using} and the UCI adult dataset \cite{kohavi1996scaling} in the following experiments.
The Pima Indian dataset consists of 768 data samples and 8 features where 6 features are continuous and integer-valued, and 2 features are continuous float-valued. The task is to predict the risk of diabetes (1: At risk, 0:Not at risk). We train a 4 layer neural network with an input layer, 2 hidden layers of 20 neurons each, and an output layer with a 80-20 training-test split. The accuracy of the model is 99.6\% on the test set. An initial population of 500 individuals is considered and the evaluation is done across 300 generations. 
The UCI adult dataset consists of 48842 samples with 14 categorical and continuous integer features and a binary outcome of predicted income ($>$50k or $<=$50k). Since the dataset contains many categorical variables, finding a counterfactual using \cite{wachter2017counterfactual} would not be feasible. We train a 6 layer neural network with an input layer, 4 hidden layers of 80 neurons each, and an output layer with a 80-20 training-test split. The accuracy of the model is 99.20\% on the test set. Since the dataset is larger, an initial random population of 1000 individuals is considered and the evaluation is done across 500 generations. The negative outcome is considered to be income $<=$50k, and we find the counterfactuals for those.
We only consider those input instances where the model prediction matches the ground-truth.

\begin{figure}[tp]
\includegraphics[width = 8.5cm,height = 5.3cm]{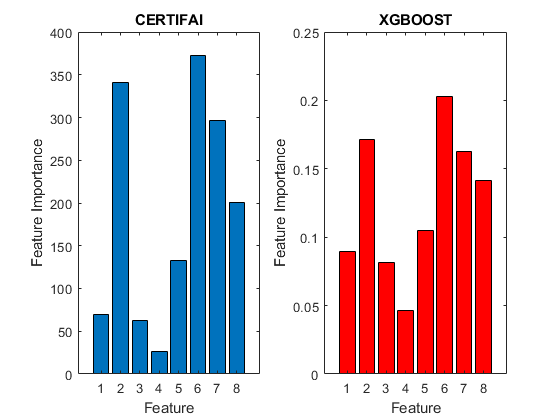}
\caption{Feature importance for the model, trained on the Pima Indian diabetes dataset, measured by the number of times a feature changed to generate the counterfactual (left) and feature importance by XGBoost (right).}
\label{fig:Pima Indian_feat_imp}
\end{figure}
\subsubsection{Importance of Constraints}
We consider two cases of counterfactual generation, counterfactuals with constraints (CWC) and counterfactuals unconstrained (CUC) for users with a prediction of high diabetes risk. CWC corresponds to a user or model creator providing a range of values for features. CUC corresponds to a user only providing the black-box model and the input instance without any constraints on the feature values. We show features for which the values have changed (between the input and counterfactual), all other values remained constant. 

As shown in Table~\ref{tab:Hi}, for person 1, when we provide constraints (CWC), the explanation is: \textit{Had your glucose been less by 34, you wouldn't have been at the risk of diabetes}. All other feature values for the user remained constant. Without constraints, the explanation shows that the BMI would have to be decreased to 10.1. While this is a smaller change in magnitude as compared to changing the glucose level, achieving a BMI of 10.1 is not feasible, and hence it is important to use the flexibility of our approach to add additional constraints that ensure feasibility. Similarly, for person 2, the age is suggested to be changed, which is not feasible. 

\subsubsection{Measuring feature importance}

From a model developer's perspective, counterfactuals can show the importance of every feature value to the prediction and hence provide transparency. If \methodName~is changing a particular feature more often than another feature when comparing the input and counterfactual, that feature is more significant for a model. For the Pima Indian diabetes dataset, we generate counterfactuals for all samples (irrespective of prediction) and analyze the number of times every feature value has changed, as shown in Figure~\ref{fig:Pima Indian_feat_imp}. Interestingly, the importances are qualitatively similar to those returned by Python's XGBoost \cite{Chen:2016:XST:2939672.2939785} library (also shown in Figure~\ref{fig:Pima Indian_feat_imp}). Specifically, feature 5 (BMI) and feature 2 (Glucose) are the most important in predicting diabetes risk. This analysis can be extended to the multi-class case by constraining sampled individuals such that they belong to a desired class (Equation~\ref{eq:Imod})
\begin{table}
\centering
\begin{tabular}{lrrrr}  
\toprule
Person & Feature & FitnessM & FitnessU\\
\midrule
1 & Race       & 0.63  & 0.87     \\
\midrule
2 & Gender       & 0.41  & 0.62     \\
\midrule
3 & Race & 0.81 & 0.81 \\
\bottomrule
\end{tabular}
\caption{Fitness values when race and gender attributes are muted (FitnessM) and unmuted (FitnessU) for three people.}
\label{tab:ind_fairness}
\end{table}
\subsubsection{Multiple counterfactual explanations}
Multiple explanations are helpful to a user so that they can receive a diverse set of changes that could be made to achieve a desired outcome. The UCI adult dataset (CWC case) is considered and features such as native-country are muted and a set range is given for features like hours-per-week. We run the genetic algorithm for the input instance and select the best two individuals that have different changes in feature indices. The advantage of our approach is that we only need to run the algorithm once, and we can generate many explanations, as opposed to \cite{russell2019efficient} where the IP solver needs to be run multiple times to generate multiple explanations. 

To underscore the benefits of suggesting alternative counterfactuals, Table~\ref{tab:boo} shows two sets of explanations that are generated by \methodName~for the same person.
Multiple explanations, the number of which is set by the user, allow a user to decide which counterfactual may be the most actionable.
\subsection{Fairness}
We evaluate fairness from an individual's perspective and from a model developer's perspective.
To see if the model is unfair towards any instance, we consider 100 random instances of the UCI adult dataset where the prediction was unfavorable and run the algorithm twice, once when the sensitive attribute is not allowed to change and once when it is, and record the fitness values. We do this for two sensitive attributes, race and gender.

The results for three such instances are shown in Table~\ref{tab:ind_fairness}. FitnessM refers to the fitness value when the race feature is muted for an individual and FitnessU corresponds to the feature being unmuted. The fitness for the first 2 people increases substantially when these protected features are allowed to change and hence for these instances, there is evidence that the model has not been fair. For the third person, the evidence suggests that the model has been fair.

\begin{figure}[tp]
\centering
\includegraphics[width = 8.5cm, height = 6cm]{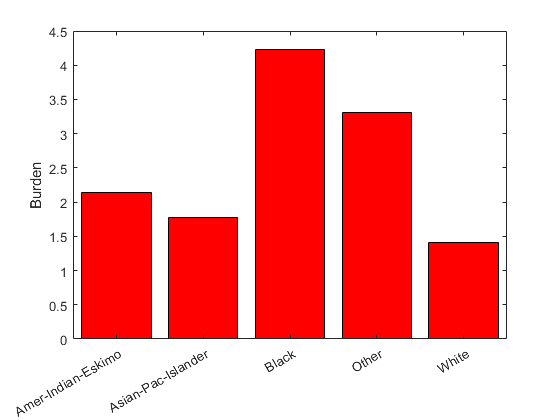}
\caption{Burden on different groups belonging to a particular race in the UCI adult dataset, found using the distance between the input instances and counterfactuals (Equation~\ref{eq:burd})}
\label{fig:burden}
\end{figure}


A model developer can use the idea of burden (Equation~\ref{eq:burd}) to evaluate how fair a model is being to groups of individuals.
To demonstrate the idea of burden, we consider the attribute race in the UCI adult dataset and take all training examples that have an unfavorable outcome. 
Results of our experiments are shown in Figure~\ref{fig:burden}. As we can see, the burden on Black race and the Other race is more than the other races. This means that on average, these groups would have to make more changes to achieve a desired prediction as compared to others. Hence the model imposes a burden on these groups, which could imply that the model has been unfair. 
\section{Conclusion and Future Work}

In this paper, we introduced \methodName, a model-agnostic, flexible, and user-friendly technique that helps build responsible artificial intelligence systems. We demonstrate the flexibility that the genetic algorithm brings to provide feasible counterfacual explanations to a user. We show how individual and group fairness can be measured using the fitness values obtained during counterfactual generation. Finally, we show how these counterfactuals are effective adversarial examples and we define CERScore, the first ever measure of robustness for a black-box model.
We are currently developing the User-Interface to CERTIFAI. Future work involves speeding up the genetic algorithm by techniques like \cite{harik1999compact} and \cite{mitchell1994will} 
A comparison between the introduced fairness metric and previous metrics would also be useful. It would also be interesting to see how our adversarial examples perform with strategies \cite{papernot2016distillation}, \cite{madry2017towards} that are aimed to handle adversarial attacks.

\bibliographystyle{named}
\bibliography{ijcai19}

\end{document}